\documentclass[a4paper,english]{rnti}

\usepackage{graphicx}

\usepackage{amssymb}
\usepackage[cmex10]{amsmath}

\usepackage{algorithm,algorithmic}

\titrecourt{Transfer Learning Using Logistic Regression in Credit Scoring}

\nomcourt{Beninel et al.}

\titre{Transfer Learning Using Logistic Regression in Credit Scoring}

\auteur{Farid Beninel\affil{1},
Waad Bouaguel\affil{2},
      Ghazi Belmufti\affil{3}}

\affiliation{
    \affil{1}CREST-ENSAI
Campus de Ker-Lann,
Rue Blaise Pascal - BP 37203,
35172 BRUZ cedex\\
          farid.beninel@ensai.fr\\

    \affil{2}LARODEC-ISG, 41 rue de la liberté, Cité Bouchoucha, 2000 Le Bardo, Tunisie\\
          bouaguelwaad@mailpost.tn,\\

    \affil{3}LARODEC-ISG, ESSEC, Université de Tunis.\\
          belmufti@yahoo.com
 }

\resume{%
The credit scoring risk management is a fast growing field due to
consumer's credit requests. Credit requests, of new and existing customers, are
often evaluated by classical discrimination rules based on customers information.
However, these kinds of strategies have serious limits and don't take into
account the characteristics difference between current customers and the future
ones. The aim of this paper is to measure credit worthiness for non customers
borrowers and to model potential risk given a heterogeneous population formed
by borrowers customers of the bank and others who are not. We hold on previous
works done in generalized gaussian discrimination and transpose them into the
logistic model to bring out efficient discrimination rules for non customers' subpopulation.
Therefore we obtain several simple models of connection between
parameters of both logistic models associated respectively to the two subpopulations.
The German credit data set is selected to experiment and to compare
these models. Experimental results show that the use of links between the two
subpopulations improve the classification accuracy for the new loan applicants.
}

\summary{%
La gestion du risque de crédit est un domaine en forte croissance, en raison du développement
des offres. Les demandes de crédit des nouveaux et anciens clients sont souvent évaluées
selon des règles de discrimination classiques basées sur le comportement, en remboursement,
des anciens clients. Ce genre dapproche présente des limites et ne tient pas compte de la
différence de caractéristiques entre les clients et les non-clients. Le but de cette étude est de
prédire le comportement des emprunteurs non-clients, étant donnée une population hétérogène
formée par les clients de la banque et les autres. Ce travail sinspire des travaux dextension de
lanalyse discriminante, par transfert de modèles gaussiens et transpose lidée au modèle logistique.
Nous nous intéressons, en particulier, à la mise en place de modèles de liaison, simples
et parcimonieux, entre paramètres des modèles logistiques associés respectivement, aux deux
sous-populations. Les différents modèles de transfert ont été expérimentés et comparés, sur la
base de données consistant en 1000 crédits à la consommation dune banque allemande.

}

\begin{document}

%
\section{Introduction}

The credit risk is one of the major risks that a loans institution has to manage. This risk arises
when a borrower doesnt pay his debt in the fixed due. To face up this kind of risk, banks' managers have to look for efficient solutions to well distinguish good from bad risk applicant. Credit scoring is one of the most successful financial risk management solutions developed for lending institutions, this solution has been fundamental in consumer credit management since \cite{Durand:1941}. Authors like \cite{Feldman:1997}, \cite{ Thomas:2002} and  \cite{Saporta:2006} defined the Credit scoring as the process of determining how likely a particular applicant is default with reimbursement.

\noindent  Credit scoring methods are applied in order to classify possible creditors in two classes of risk: good and bad \citep{GIUDICI:2003}. These methods use explanatory variables obtained from applicant information to estimate his intended performance to pay back loan.  A large number of Transfer Learning Using Logistic Regression in Credit Scoring classification methods can be used in the process of identifying borrowers behavior as decision trees \citep{Breiman:1985}, neural networks \citep{Mcculloch:1943}, discriminant analysis \citep{Fisher:1936, Mahalanobis:1936}, logistic regression \citep{Cox:1970,Cox:1989} \ldots

    \noindent Both these techniques can provide good discrimination but the most common used methods for building scorecard (i.e. credit models) are discriminant analysis and logistic regression. Logistic regression is a more appropriate technique for Credit scoring cases \citep{Henley:1996}. \cite{Fan:1998} and \cite{Sautory-Chang:1992} recommend the use of the binary logistic regression in Credit scoring cases, when discriminant analysis application conditions are not obtainable. This choice becomes imperative if qualitative variables get involved in the model \citep{Bardos:2001}.

\noindent Available information about credit candidate supplies a fundamental element in his credit request acceptation, information lack in credit risk valorization is suspected to lead to wrong decision making. In this paper, we will focus in Credit scoring evaluation, using logistic regression technique when the population of interest is characterized by a small size.

\noindent  Borrower's behavior is described by a binary target variable denoted $Y$, value taken by this last one supplies a basic element in credits' granting decision, $Y =0$ when the borrower presents problem and $Y = 1$ otherwise. Beside this variable, every borrower is also described by a set of description variables $(X_1,X_2,\ldots,X_d)$ informing about the borrower and about his accounts' functioning.

 \noindent The sample of loans' applicants results from a heterogeneous population formed by borrowers customers and others who are not. Here we deal with the problem of discrimination in the case of a subpopulations' mixture, where the two subpopulations are respectively:  borrowers' customers and borrowers' non customers. More precisely, we will focus in non customers subpopulation credit worthiness evaluation,  assuming that sample size of this subpopulation is  considered weak.

 \noindent Beginning with the hypothesis that population size is one of the most important factors affecting
the classification power of the logistic regression technique, we evaluate future customers
(i.e non customer) behavior to pay back loan, by looking for efficient solution to the problem
of non customers small sample size.

 \noindent We proceed to investigate how using the information on hand of borrowers customers and non
customers can be efficient. The first approach, which is generally used by banks, consists in
using the borrowers customers predictive model, to predict borrowers non customers behavior.
However, it does not take into account, difference between the two subpopulations. Another
approach consists in using a learning sample resulting from non customers subpopulation to
build their predictive model. However, this second approach needs a learning sample of a suitable
size, which is not our case.

Changing the second approach can bring an efficient solution to the problem of learning from
small size sample. This change consists in using a design sample, drown from another population considered slightly different (e.g. customers subpopulation), this sample will be used in
models building in place of non customers design sample. The idea of using two slightly populations
for estimating one population parameters, has been first proposed by \cite{Biernacki:2002}. In a multinomial context, \cite{Biernacki:2002} proved that two slightly different populations
are linked through linear relations. Estimation of nonlabeled sample allocation rules
was obtained via estimating the linear relationship parameters, using five constraints models
on the linear relationships.

\noindent This approach proved to be efficient in biological context and many extension of this paper was proposed, including \cite{Biernacki-Jacques:2007}, \cite{Bouveyron-Jacques:2009} as well as \cite{Beninel-Biernacki:2005,Beninel-Biernacki:2007,Beninel-Biernacki:2009}. \cite{Beninel-Biernacki:2009} extended this approach to the multinomial logistic discrimination and proposed an additional links' model in the case where the two studied subpopulation are two Gaussian ones. The main idea of this previous works is that information related to one of the two subpopulations contains some information related to the other one.

\noindent The earlier works, have been the exit of the main thoughts of this paper, given previous results in the case of Gaussian mixture model and the six presented models by \cite{Beninel-Biernacki:2009}, our task is then, to go in deep in the previous results with more testes and simulations, and to add to the former links' models a seventh one.

\section{Logistic Regression Model}
\subsection{Classical Logistic Regression}
Logistic regression is a variant of linear regression, which is used when the dependent variable is a binary variable and the independent variables are continuous, categorical, or both. Logistic regression model supplies a linear function of descriptors as discrimination tool, this technique is widely used in Credit scoring applications due to its simplicity and explainability. Model form is given by
\begin{center}
\begin{equation}
log(\displaystyle \frac{p_i}{1-p_i})=\beta_0 +\beta^T \mathbf{x}_i,
\end{equation}
\end{center}

where
\begin{itemize}
\item $p_i$ is the posteriori probability, defined as the probability that an individual $i$ have the modality $1$ for given values taken by descriptors (i.e. $P(Y_i=1|\mathbf{x}_i)$).
\item $\mathbf{x}_i = (\mathbf{x}_i^1, \mathbf{x}_i^2,\ldots, \mathbf{x}_i^d)$  is the vector of observed value taking by description variables.
 \item $\beta^T=(\beta_1,\beta_2,\ldots,\beta_d)$ is the vector of variables effect.
\item $\beta_0$ is the intercept.
\end{itemize}

\noindent This technique serves to estimate the posteriori probability $p_i$, which the value allows to assign every borrower to his group membership i.e., $\{Y_i=1\}$ $if$ $p_i$ is greater than a fixed threshold
value and $\{Y_i=0\}$ otherwise. 


\subsection{Mixture Logistic Regression}
Let us remind that we deal with the problem of discrimination in case of subpopulations' mixture, where the two subpopulations of interests are the subpopulation of borrowers customers and the subpopulation of borrowers non customers, denoted respectively $\Omega$ and $\Omega^*$, for  which we associate the two following posteriori probability $p$ and $p^*$. Our purpose is the prediction of the solvency of  borrowers' non customers using the information on hand of the two subpopulations.

Given two learning samples $S_A =\{(\mathbf{x}_i, Y_i): i = 1,..., n\}$ and $S_A^* = \{(\mathbf{x}^*_ i, Y^*_ i): i = 1,... ,n^*\} $, where the pairs $(\mathbf{x}_i, Y_i)$  and $(\mathbf{x}^*_i, Y^*_i)$ are independent and identically distributed (i.i.d.) realizations of the random couples $ (\mathbf{x}, Y) $  and $ (\mathbf{x}^*, Y^*)$, we consider the logistic model over $\Omega$, as given by
\begin{center}
\begin{equation}
p_i=P(Y_i= 1| \mathbf{x}_i,\theta)= \displaystyle \frac{exp^{\beta_0 +\beta ^T \mathbf{x}_i}}{1+exp^{\beta_0 +\beta ^T \mathbf{x}_i}},
\end{equation}\label{score1}
\end{center}

and over  $\Omega^*$
\begin{center}
\begin{equation}
p^*_i= P (Y_i^*= 1| \mathbf{x}_i^*,\theta^*)= \displaystyle \frac{exp^{\beta^*_0 +\beta^{*T} \mathbf{x}_i^*}}{1+exp^{\beta^*_0 +\beta^{*T} \mathbf{x}_i^*}},
\end{equation}\label{score2}
\end{center}
where
\begin{itemize}
\item $\theta= \{( \beta_0|| \beta^T) \in \mathbb{R}^{d+1}\}$  and  $\theta^*  = \{( \beta_0^*|| \beta^{*T}) \in \mathbb{R}^{d+1}\}$ are the sets of all parameters to be estimated respectively over  $\Omega$ and $\Omega^*$.
\item $( \beta_0|| \beta^T) $ and $( \beta_0^*|| \beta^{*T}) $ are the concatenations of the intercept  and the vector of variables  effect over   $\Omega$ and $\Omega^*$.
\end{itemize}
The mixture model allows the resolution of various discrimination problems, in our case we assume that an experienced rule, to predict on the first subpopulation $\Omega$ is known and we have a small learning sample from the second subpopulations $\Omega^*$. From available data we want to get a new allocation rule over $\Omega^*$.

\noindent According to \cite{Beninel-Biernacki:2009} links between subpopulations could exist and consequently, information on $\Omega$ could provide some information on $\Omega^*$. Existence of a link between variables vector implies a link between the two scores functions given in (\ref{score1}) and (\ref{score2}). Using acceptable links between the scores functions of the two subpopulations allows to use hidden information of samples $S^*_L$ and $S_L $ to get the allocations rules over $\Omega^*$. We look in what follows for these links basing on results found in Gaussian case.

\section{Gaussian Case and Links Models}
In order to estimate the score function parameters over $\Omega^*$, we use the data on hand of the two subpopulations. The use of customer subpopulation $\Omega$ data aims to moderate the small size of the subpopulation $\Omega^*$ of non customers, by supposing the existence of hidden links between the distribution of variables over $\Omega$ and that over $\Omega^*$.

\noindent It's known from \cite{Beninel-Biernacki:2007} as well as \cite{Bouveyron-Jacques:2009} that existence of particular connections between the variables distributions lead to relations between the parameters of their respective logistic regression models, consequently our task consists in finding these links. In this context a preliminary case study was successfully done in Gaussian multivariate case \cite{Beninel-Biernacki:2005}. It is a question here of extending the found results in Gaussian case to logistic case, which leads to simple and parsimonious linking models between the parameters of logistic classification rules associated respectively to the two subpopulations $\Omega$ and $\Omega^*$.

\subsection{Gaussian Case: Subpopulations Links}
In Gaussian discrimination, it is crucial to define handled data in terms of two samples: a learning sample $L$ and a prediction sample $P$, resulting respectively from the following subpopulations: $\Omega$ and $\Omega^*$. In our case these two subpopulations are different.

The learning sample $L$ is composed of $n$ pairs $ (\mathbf{x}_i, Y_i) $, $i = 1, \ldots, n$ where, $\mathbf{x}_i$ is a vector of  $\mathbb { R }^d$  representing the numeric characteristics describing the individual $i$ and $Y_i$  is  his group's label. The $n$ pairs $ (\mathbf{x}_i, Y_i) $ are supposed to be i.i.d realizations of the random couple $ (\mathbf{x}, Y) $ defined over $\Omega$ by the following  joint distribution:

\begin{center}
\begin{equation}
\begin{tabular}{lll}
$\mathbf{x}_{|Y=k}\sim$\emph{$N_d$}$(\mu_k, \Sigma_k)$ $k= \{1,..., K\}$\\
&and\\
$Y\sim$\emph{$M_K$}$(1, \pi_1,...,\pi_K)$,
\end{tabular}
\end{equation}
\end{center}
where \emph{$N_d$}$(\mu_k, \Sigma_k)$ is the Gaussians distributions of dimension $d$, with an average $\mu_k$ and a variance-covariance matrix $\Sigma_k$. \emph{$M_K$}$(1, \pi_1,...,\pi_K)$ is the multinomial distribution  of parameters $\pi_1,...,\pi_K$, where  $\pi_k$ is  the proportion of the group $k$ in the subpopulation and the parameter  $K$ represents modality of the target variable $Y$.

The prediction sample $P$ consists of $ n^*$ individuals, which we know their numeric characteristics $\mathbf{x}^*_i$, $i = 1,...,n^* $, assumed the same over $L$. The $n^*$ labels $Y ^ *_ 1, \ldots,Y^ * _ {n ^ *} $ are to be estimated. The $n^*$ pairs $ (\mathbf{x}^*_i, Y^*_i) $ are supposed to be i.i.d realizations of the random couple $ (\mathbf{x}, Y) $ defined over $\Omega^*$ by the following joint distribution:

\begin{center}
\begin{equation}
\begin{tabular}{lll}
$\mathbf{x}^*_{|Y^*=k}\sim$\emph{$N_d$}$(\mu^*_k, \Sigma^*_k)$,  $k= \{1,..., K\}$\\
&and\\
$Y^*\sim$\emph{$M_K$}$(1, \pi^*_1,...,\pi^*_K)$.
\end{tabular}
\end{equation}
\end{center}

Then, we try to estimate the $n^*$ unknown labels by using  resulting information from the samples $L$ and $P$. Our task is then, to identify acceptable relations linking the two subpopulations. In order to bring to light the existing  links between the two subpopulations, we are going to adopt the approach proposed by Beninel and Biernacki \cite{Beninel-Biernacki:2009}, which supposes the existence of an application $\phi_k : \mathbb{R}^{d} \rightarrow \mathbb{R}^{d}$ linking in law the random variables vectors of $\Omega$ and $\Omega^*$. Then

\begin{center}
\begin{equation}\label{fi}
\begin{tabular}{lll}
$ \mathbf{x}^*_{|Y^*=k}\sim \phi_k (\mathbf{x}_{|Y=k})=[\phi_{k1} (\mathbf{x}_{|Y=k}),...,\phi_{kd} (\mathbf{x}_{|Y=k}) ]^T$.
\end{tabular}
\end{equation}
\end{center}

The outcomes resulting from \cite{Beninel-Biernacki:2009} verify that the function $\phi _ k$ is affine, we drive from equation (\ref{fi}) the following relations between the variables distributions:
\begin{center}
\begin{equation}
\mathbf{x}^*_{|Y^*=k}\sim \Lambda_k \mathbf{x}_{|Y=k}+\alpha_k,
\end{equation}
\end{center}
where $\Lambda_k$ is a diagonal matrix defined over $\mathbb{R}^{d\times d}$ and $\alpha_k$ is a vector of $\mathbb{R}^d$. From the previous expression we deduct this following  links between the parameters of two subpopulations

\begin{alignat}{2}
\label{lien1}
\mu^*_k = \Lambda_k \mu_k+\alpha_k, \\ \label{lien2}
\Sigma^*_k = \Lambda_k \Sigma_k \Lambda_k. \\
\notag 
\end{alignat}

\subsection{Gaussian Case Extended to Logistic Case}
\cite{Anderson:1982} proved the existence of a link between the parameters of the mixture Gaussian model and those of corresponding logistic model. Links between the two subpopulations can be obtained in a stochastic case where, the variables vector $\mathbf{x}$ and $ \mathbf{x}^*$ defined over $\Omega$ and $\Omega^*$ are Gaussian, homoscedastic conditionally in the groups and the matrices of common variance-covariance are noted in the following way:
\begin{center}
\begin{equation}
\Sigma = \Sigma_1 = \Sigma_2 and  \Sigma^* = \Sigma^*_1 = \Sigma^*_2,
\end{equation}
\end{center}

we obtain the following links between the logistic parameters and the Gaussian one for the two subpopulations:

\noindent over $\Omega$,
\begin{center}
\begin{equation}
\beta_0 = \frac{1}{2}(\mu_2^T \Sigma^{-1}\mu_2 -\mu_1^T \Sigma^{-1}\mu_1) and  \beta =\Sigma^{-1}(\mu_1 -\mu_2)
\end{equation}

\end{center}
and over $\Omega^*$,

\begin{center}
\begin{equation}
\begin{tabular}{ll}
$\beta^*_0 = \frac{1}{2}(\mu_2^{*T} \Sigma^{*-1}\mu_2^* -\mu_1^{*T} \Sigma^{*-1}\mu_1^*)$ and $ \beta^* =\Sigma^{*-1}(\mu_1^* -\mu_2^*) $
\end{tabular}
\end{equation}
\end{center}

\noindent replacing the $\mu_k^{*}$, $k=1,2$ and the $\Sigma^{*}$ by their expression given by equations (\ref{lien1}), (\ref{lien2}) and  limiting to linear relations which, can exist between the two subpopulations parameters, we obtain the following expressions  for $\beta_0^*$ and $\beta^*$:
\begin{center}
\begin{equation}\label{eq13}
\begin{tabular}{lll}
$\beta^*_0 =c + \beta_0$ and $ \beta^* =\Lambda \beta $,
\end{tabular}
\end{equation}
\end{center}
consequently, the scoring function obtained by replacing the parameters $\beta^*_0$ and $\beta^*$ in equation (\ref{score2}) is given by:
\begin{center}
\begin{equation}\label{eq14}
\begin{tabular}{lll}
$ P (Y_i^*= 1| \mathbf{x}_i^*,\theta,\varrho)=\displaystyle \frac{exp^{\beta_0 +c+(\Lambda\beta)^T \mathbf{x}_i^*}}{1+exp^{\beta_0 +c+(\Lambda\beta)^T \mathbf{x}_i^*}}$,
\end{tabular}
\end{equation}
\end{center}

\noindent here $\varrho=\{(c,\Lambda)\in \mathbb{R}^{d+1}\}$ is the set of transition parameters to be estimated.
\subsection{Links Models}
Estimation of links between $\Omega$ and $\Omega^* $ subpopulations is done through several logistic intermediary sample models of connections, inspired by the Gaussian case previously evoked in subsection 3.1. Our purpose in this paper is the estimation and comparison of this models listed in the following table

\begin{table}[ht]
 \begin{center}
   \tabcolsep = 2\tabcolsep
   \begin{tabular}
{|c|cl|p{10cm}|}
\hline
\textbf{Models}& \multicolumn{2}{|c|}{\textbf{Parameters}}&\textbf{Descreptions}\\\hline
$M1$& $c= 0$&$\Lambda =I_d$& The score functions are invariable.\vspace{3mm}\\ \hline
 $M2$& $c = 0$&$\Lambda =\lambda I_d$& The score functions of the two subpopulations differ only through the scalar parameter $\lambda$.\\ \hline
$M3$& $c\in \mathbb{R}$&$\Lambda =I_d$& The score functions of the two subpopulations differ only through the scalar parameter  $\beta^*_0$.\\ \hline
$M4$& $c \in \mathbb{R}$&$\Lambda =\lambda I_d$& The score function of the two subpopulations differ through the couple ($\beta^*_0$, $\lambda$).\\ \hline
$M5$& $c = 0$&$\Lambda \in \mathbb{R}^{d\times d}$& The score functions of the two subpopulations differ only through the vectoriel parameter  $\beta^*$.\\ \hline
 $M6$& $c \in \mathbb{R}$&$\Lambda \in \mathbb{R}^{d\times d}$& There is no more stochastic link between the logistic discriminations of the two subpopulations. All parameters are free. \\ \hline
\end{tabular}
\caption{Links models}\label{table1}
\end{center}
\end{table}

For each one of the above models, estimation of transition parameters is conditionally done to  the subpopulation $\Omega$ parameters. We add a seventh model noted $M7$, which consist in introducing as observations, all the borrowers (customer and non customers) and to apply a simple logistic regression. This consists in the joined estimation of $\Omega$ parameters and  the transition parameters.
\section{Empirical Analysis}
\subsection{Credit Data Set and Subpopulations Definition}
\begin{figure}[t]
\begin{center}
 \includegraphics[width=6cm]{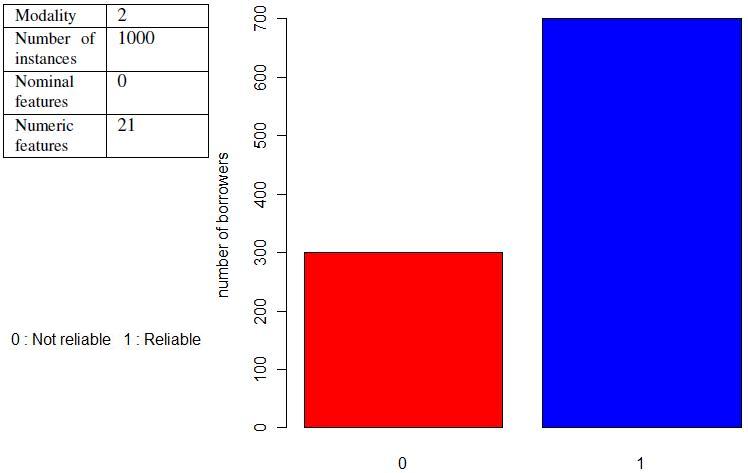}
  \caption{German data set description. }\label{freq1}
\end{center}
\end{figure}

The adopted herein data set is a real word data set: German credit data,
 illustrated in Figure \ref{freq1}, available from the UCI Machine Learning Repository (http://archive.ics.uci.edu/ml/datasets.html) or see also \citep{citeulike:155081} for more description. The German Credit scoring data set is often used by credit specialists. It cover a sample of $1000$ credit consumers where $700$ instances are creditworthy applicants and $300$ are not. Each applicant is described by a binary target variable $Kredit$,  $Kredit=1$ for creditworthy  and $Kredit=0$ otherwise,
   $20$ other input variables are assumed to influence this target variable, duration of credits in months ($Laufzeit$), behaviour repayment of other loans ($Moral$),
   value of savings or stocks ($Sparkont$), stability in the employment ($Beszeit$), further running credits ($Weitkred$) \ldots

In this case study we are interested in the evaluation of the borrowers non customers behavior to pay back loans, we use the variable $Laufkont$ (balance of current account) to separate the available  data set in two subpopulations: $ Laufkont >1$ the customers subpopulation composed from $726$,   $ Laufkont =1$ the non customers subpopulation composed from $274$.

\noindent Afterward, we devine the subpopulation of  borrowers non customers into two samples: a learning sample $S_L^*$ and a test sample $S_T^*$. The first sample allows to represent the diverse models and to bring out affectation rules, the second one allows to verify the reliability of the established models in learning step.

\subsection{Experiments Description}
To obtain a robust estimate of our seven models performance, our simulations involves taking $50$ random design (of size $n\in\{50,100,150,200\}$) and test sample splits from the non customers subpopulation. For each design the following algorithm is applied to estimate the parameters of each model from our seven logistic models.
\begin{algorithm}[htp]
\caption{\sl $ESTIM (\mathbf{x},\mathbf{x}^* , S_L^*)$ $\rightarrow \theta^*$}
	\begin{algorithmic}[1]

\REQUIRE
 $\mathbf{x}$: Customer design matrix defined over $\Omega$.\\
  $\mathbf{x}^*$: non customer design matrix defined over $\Omega^*$.\\
 $S_L^*$: Non customer learning sample, $S_L^*\in \Omega^*$.\\
\ENSURE
 $\theta^*$:  Set of all parameters to be estimated over  $\Omega^*$, $\theta^*  = \{( \beta_0^*|| \beta^{*T}) \in \mathbb{R}^{d+1}\}$.
\STATE Estimate the set of parameters $\theta$, $\theta= \{( \beta_0|| \beta^T) \in \mathbb{R}^{d+1}\}$,  using a simple logistic regression on $\mathbf{x}$.
\STATE  Estimate the set of transition parameters $\varrho$, $\varrho=\{(c||\Lambda)\in \mathbb{R}^{d+1}\}$ using the learning sample $S_L^*$ and the design matrix $\mathbf{x}^*$.
\STATE Replace the parameters in equation (\ref{eq13}) by their values found  in Step 1 and Step 2.
\STATE{ return $\theta^*$.}
\end{algorithmic}
\end{algorithm}

Table II summarizes the different parameters to be estimated by the previous algorithm for the seven studied models. Once all parameters are estimated, an estimate of the applicant group label is obtained by replacing the parameters by their values in equation (\ref{eq14}).

Most application of assignment procedures works with the misclassification error rates as evaluation criterion, in this case study our choice was a unusual one, so we decided to work with the test error rate, Type $II$ error rate and  Type $I$ error rate. The aim was to focus in minimizing the number of default accepted applicant by minimizing the Type $I$ error.

\begin{table}[ht]
 \begin{center}
   \tabcolsep = 2\tabcolsep
   \begin{tabular}{|ll|p{3cm}|p{4cm}|l|}
  \hline

 \multicolumn{2}{|c|}{ \textbf{Models}}&\textbf{Learning sample}& \textbf{transition parameters }& \multicolumn{1}{c|}{\textbf{Estimated parameters}} \\
  \hline
  $M1$& & $\Omega$& &$\hat{\beta^*_0} =\beta_0$ and $ \hat{\beta^*}= \beta $  \\
 \hline
  $M2$& & & $\lambda$& $\hat{\beta^*_0} = \beta_0$ and $ \hat{\beta^*} =\lambda  \beta $\\
  \cline{1-2}
  \cline{4-5}

   $M3$&  &  &$c$ &$\hat{\beta^*_0} =c + \beta_0$ and $ \hat{\beta^*} =\beta $\\

  \cline{1-2}
  \cline{4-5}

    $M4$&  & $S_L^*$&$c$ and $\lambda$&$\hat{\beta^*_0 }=c + \beta_0$ and $\hat{\beta^*} =\lambda \beta $ \\
\cline{1-2}
  \cline{4-5}

   $M5$& & &$\Lambda$&$\hat{\beta^*_0 }= \beta_0$ and $\hat{ \beta^*} =\Lambda \beta$ \\

\cline{1-2}
  \cline{4-5}
  $M6$&  &  & $c$ and $\Lambda$ & $\hat{\beta^*_0} =c + \beta_0$ and $ \hat{\beta^*} =\Lambda \beta $\\
\hline
     $M7$& &  $S_L^*\cup \Omega$  &  & $\hat{\beta^*_0} $ and $\hat{\beta^*}$\\
\hline
\end{tabular}
\label{tab0}

  \caption{ Summary of  parameters to be estimated}

\end{center}
\end{table}

\subsection{Experimental Results}
The results for the German credit data set were obtained by using the seven models are summarized in Tables \ref {tab1},  \ref{tab2} and \ref {tab3}  respectively.

\begin{table}[ht]
 \begin{center}
   \tabcolsep = 2\tabcolsep
   \begin{tabular}{lp{0.45cm}p{0.45cm}p{0.45cm}p{0.45cm}p{0.45cm}p{0.45cm}p{0.45cm}}
\hline
\multicolumn{1}{c}{ \textbf{Models}}&$M1$&$M2$&$M3$&$M4$&$M5$&$M6$&$M7$\\
\hline
$n=50$&$0.348
$&$0.370
$&$0.348
$&$0.347
$&$0.358
$&$0.361
$&$0.343
$\\
\hline
$n=100$&$0.385
$&$0.362
$&$0.344
$&$0.345
$&$0.347
$&$0.344
$&$0.385
$\\
\hline

$n=150
$&$0.356
$&$0.354
$&$0.330
$&$0.332
$&$0.338
$&$0.342
$&$0.354
$\\
\hline
$n=200$&$0.367
$&$0.337
$&$0.308
$&$0.308
$&$0.321
$&$0.315
$&$0.334
$\\
\hline

\end{tabular}
\caption{ Results summary for predictive credit  test error rate with respect to the learning sample size}\label{tab1}
\end{center}
\end{table}
\begin{table}[ht]
 \begin{center}
   \tabcolsep = 2\tabcolsep
   \begin{tabular}{lp{0.45cm}p{0.45cm}p{0.45cm}p{0.45cm}p{0.45cm}p{0.45cm}p{0.45cm}}
\hline
\multicolumn{1}{c}{\textbf{Models}}&$M1$&$M2$&$M3$&$M4$&$M5$&$M6$&$M7$\\
\hline
$n=50$&$0.282
$&$0.312
$&$0.338
$&$0.332
$&$0.385
$&$0.341
$&$0.283
$\\
\hline
$n=100$&$0.226
$&$0.304
$&$0.339
$&$0.311
$&$0.364
$&$0.356
$&$0.284
$\\
\hline
$n=150$&$0.209
$&$0.286
$&$0.296
$&$0.321
$&$0.344
$&$0.338
$&$0.279$
\\
\hline
$n=200$&$0.185
$&$0.283
$&$0.294
$&$0.296
$&$0.305
$&$0.245
$&$0.203
$\\
\hline
\end{tabular}
  \caption{ Results summary for predictive credit Type $II$ error with respect to the learning sample size}
\label{tab2}
\end{center}
\end{table}

\begin{table}[ht]
 \begin{center}
   \tabcolsep = 2\tabcolsep
   \begin{tabular}{lp{0.45cm}p{0.45cm}p{0.45cm}p{0.45cm}p{0.45cm}p{0.45cm}p{0.45cm}}
\hline
\multicolumn{1}{c}{ \textbf{Models}}&$M1$&$M2$&$M3$&$M4$&$M5$&$M6$&$M7$\\
\hline
$n=50$&$0.394
$&$0.321
$&$0.284
$&$0.285
$&$0.291
$&$0.301
$&$0.376
$\\
\hline
$n=100$&$0.384
$&$0.301
$&$0.279
$&$0.275
$&$0.281
$&$0.297
$&$0.362
$\\
\hline

$n=150$&$0.384
$&$0.281
$&$0.230
$&$0.233
$&$0.253
$&$0.275
$&$0.316
$\\
\hline
$n=200$&$0.336
$&$0.271
$&$0.220
$&$0.218
$&$0.250
$&$0.273
$&$0.278
$\\
\hline

  \end{tabular}
\caption{Results summary for predictive credit Type $I$ error with respect to the learning sample size} \label{tab3}
 \end{center}
\end{table}

We found no significant differences among models $M3$ and $M4$ that means that these two models achieved almost the same test error rate  in Table \ref {tab1} and almost the same Type $II$ and Type $I$ error in Tables \ref{tab2} and \ref{tab3}, for different training size. It is obvious from Table \ref{tab1}, that test error rate decreases proportionally to the learning sample size, this improvement can be suitable to the estimate of models' parameters which become more precise with the increase of the training data size.  Tables \ref{tab2} and \ref{tab3}, shows that Type $II$ error and Type $I$ proportionally decrease to the design sample size, these results prove the importance of the population size in classification.

As shown in Table \ref{tab1}, the test error rate of the two previous models achieved $0.308$ which is the lowest rate of misclassified instances, according to this first criterion these models are the two best classification models. For the remaining models, we remark that models $M5$ and $M6$ also achieved good results, followed by model $M2$, the left behind two models generate the most raised test error rate, specially model $M1$ which appears the worst one.

Test error rate, however measured, is only one aspect of performance, this criterion may not be the most precise one, further   misclassification rate can be another aspect of performance, so each model is evaluated by assessing Type $I$ and $II$ error rate. We remind that the cut-off threshold used in this case study is $0.5$ for this threshold, all the applicants whose estimated probability of non-reliability $P(Y = 0)$ is less  than $0.5$ are assessed as non-reliable applicants, otherwise they are classified as reliable. In Table \ref{tab2} model $M1$ and $M7$ achieved $0.185$ and $0.203$ error rate, which are the lowest Type $II$ error rate, in other hand models $M5$ and $M6$, followed by $M2$ have the most raised rate, this kind of error arise when a reliable applicant is predicted as non-reliable.  Models $M5$ and $M6$ are less efficient in the reliable applicants prediction.

\noindent  Table \ref{tab3} summarize  Type $I$ error  for the seven models. A  Type $I$ error means taking a non-reliable client and predicting him as reliable, this kind of error is more dangerous and more costly than the previous one, the model with the lowest rate of Type $I$ error is considered as the best model. From Table \ref{tab3} we remark that models $M3$ and $M4$ have the lowest rate of Type $I$ error, followed by models $M5$, $M6$ and $M2$, in other hand models $M1$ and $M7$ have the most raised Type $I$ error rate, It seems that these two previous models have greater difficulty in predicting non-reliable clients than reliable ones.

The previous misclassification rates are obtained when the cut-off is $0.5$, however changing this threshold might modify the previous results and can allow decider to catch a greater number of good or bad applicants. Hand \cite{hand:2001} in his work proposed the use of graphical tools as evaluation criterion, in place of scalar criterion. We use in this paper the ROC (i.e. receiver operating characteristic)  curve to evaluate our seven models, the ROC curve shows how the errors change when the threshold varies, this kind of curve situate positives instances against the negatives instances  which allow finding the middle ground between specificity and sensitivity.

\begin{figure}[t]
\begin{center}
 \includegraphics[width=6cm]{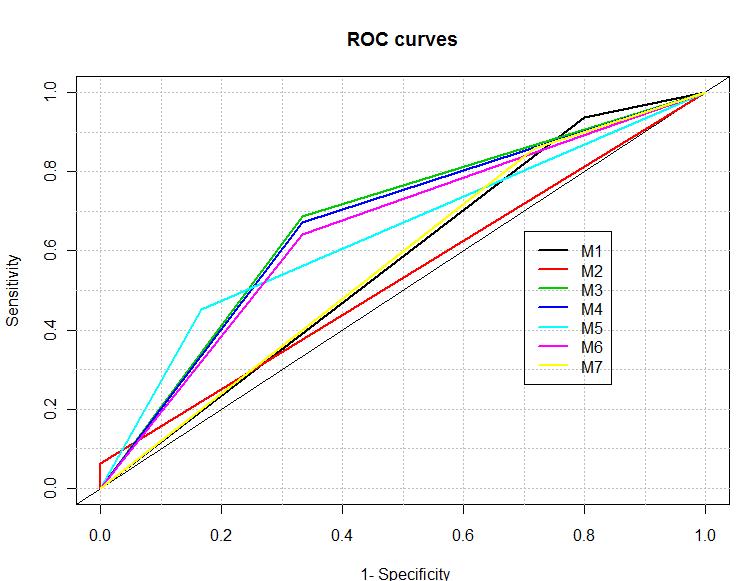}
 \caption{ROC curves. }\label{fig2}
\end{center}
\end{figure}

\noindent Figure \ref{fig2} shows the ROC curve of our models. The $X$ axis of the curve represents models' $1-specificity$ (i.e. Type $II$ error rate) and the $Y$ axis represents models' $sensitivity$ (i.e. $1-$Type $I$ error rate). According to Liu and Schumann \cite{Liu:2002} a model with a ROC curve, which follows the $45^\circ$ line would be useless. It would classify the same proportion of not worthy applicants and worthy cases into the not worthy class at each value of the threshold.  Figure \ref{fig2} shows that the seven models are convexes and situated over the first bisector, which lead us to affirm that our models are statistically approved and not useless. In Figure \ref{fig2} we remark that models $M3$ and $M4$ curves appears considerably higher to the other models' curves which confirms our intuition about their performance, models $M1$, $M7$ and $M2$ has the lowest AUC (i.e. air under curve), from the balance between false positive and false negative point of view these models are bad.

The evoked performance measures in this section, served to evaluate the validity and the discriminant power of the studied models. From the previous results, we remark that the most banks' practiced model $M1$ seem the least successful model once applied to non customers borrowers' data, this confirms the difference between the two studied subpopulations. We also remark that models $M5$ and $M6$ might be a good classifier. However, models $M3$ and $M4$ seems to be the more suitable models for the prediction of non customers behavior to pay back loan. These last one are the best predictive models because their constant is calculated from the non customers learning sample independently of customers sample, what supposes their  importance in the reliability  prediction  of the target variable $kredit$ and confirm the existence of a certain link between the two subpopulation $\Omega$ and $\Omega^*$.  Model $M5$ possesses the most raised rate of Type $II$ error, this model is considered as careful but its use can lead to a loss of reliable borrowers.

To be sure of our models performance, we compare in what follows the performance of the models $M3$ and $M4$ with two successful classification techniques:
\begin{itemize}

\item \textbf{SVM} \citep{Vapnik:1995}  is one of the most outstanding machine learning techniques. The use of SVM in financial application has been previously discussed by several works \citep{2005_SchebeschStecking, Min:2005, Huang:2006, Wang:2008,BellottiC09}. There many raisons for choosing SVM \citep{ Burges:1998},  it requires less prior assumptions about the input data   and can perform on small or huge data set by doing a nonlinear mapping from an original input space into a high dimensional feature space.
\item\textbf{Decision trees} \citep{Breiman:1985}  was used in credit scoring for the first time by \cite{Frydman:1985}.  DT is a very simple method and can be described as a set of nodes and edges, the root node define the first split of the credit-applicants sample. Each internal node   split the set of instances into two subsets. Each node contains individuals of a single class; the operation is repeated until the separation in sub-populations is no more possible.
\end{itemize}

\begin{table}[ht]
 \begin{center}
   \tabcolsep = 2\tabcolsep
   \begin{tabular}{lllll}
\hline
 $n=200$& && & \\\hline
  \textbf{Method} & \textbf{Model}&\textbf{Test Error}&\textbf{Type $I$}&\textbf{Type $II$}\\
\hline

\hline
\textbf{Logistic} &$M3$&$0.308$&$0.220$&$0.294$\\
 &$M4$&$0.308$&$0.218$&$0.296$\\

 \textbf{SVM} &Radial&$0.215$&$0.240$&$0.304$\\
 &Polynomial &$0.355$&$0.344$&$0.363$\\

  \textbf{DT}&$ID3$&$0.395$&$0.222$&$0.431$\\
 &$C4.5$&$0.275$&$0.258$&$0.298$\\
\hline
  \end{tabular}
\caption{Results summary of error rate for SVM, DT and links model M3 and M4.}\label{tab4}
 \end{center}
\end{table}

Table \ref{tab4} summarizes the average error rates using two SVM: radial and polynomial; two DT: ID3 and C4.5 and the previous result of models $M3$ and $M4$ based on a training sample of $200$ instances. Table \ref{tab4} shows that for test error rate radial SVM and C4.5 seems the best ones followed by $M3$ and $M4$. Although the performances of the added techniques, model $M3$ and $M4$ yield slightly the lower type $I$ and type $II$ error rate. The results confirm the performance of the proposed links model.

\section{Conclusion}

In this paper we have considered the problem of credit worthiness evaluation, for a population of insufficient size. We proposed   seven simple logistic submodels combining the classification rule on customers subpopulation and the labeled sample from the non customers subpopulation. A comparison of the seven models performance was done and the models $M3$ and $M4$ was selected as the best classification model for the non customers subpopulation, this two models beat  the performance of traditional classification model $M1$.

\noindent This research would have been able to generate more interesting results if we were able to have a non customers' sample of bigger size. We envisage as perspective, to apply logistic regression using non-linear links between the two subpopulations. We also can apply a non-parametric approach which can seem efficient once the linear models find their limits.

\bibliographystyle{rnti}

\bibliography{BIB}


\end{document}